\documentclass{article} 
\usepackage{iclr2026_conference,times}

\usepackage{amsmath,amsfonts,bm}

\def\eqref#1{equation~\ref{#1}}

\def\1{\bm{1}}

\DeclareMathAlphabet{\mathsfit}{\encodingdefault}{\sfdefault}{m}{sl}
\SetMathAlphabet{\mathsfit}{bold}{\encodingdefault}{\sfdefault}{bx}{n}

\usepackage{hyperref}
\usepackage{url}
\usepackage{tikz}
\usepackage{threeparttable}
\usepackage{float}
\usetikzlibrary{shapes,arrows,positioning}

\title{Evaluating LLM Simulators as Differentially Private Data Generators}

\author{Nassima M. Bouzid, Dehao Yuan, Nam H. Nguyen, Mayana Pereira \\
Capital One \\
}

\iclrfinalcopy 
\begin{document}

\maketitle

\begin{abstract}
LLM-based simulators offer a promising path for generating complex synthetic data where traditional differentially private (DP) methods struggle with high-dimensional user profiles. But can LLMs faithfully reproduce statistical distributions from DP-protected inputs? We evaluate this using PersonaLedger, an agentic financial simulator, seeded with DP synthetic personas derived from real user statistics. We find that PersonaLedger achieves promising fraud detection utility (AUC 0.70 at $\epsilon=1$) but exhibits significant distribution drift due to systematic LLM biases---learned priors overriding input statistics for temporal and demographic features. These failure modes must be addressed before LLM-based methods can handle the richer user representations where they might otherwise excel.
\end{abstract}

\section{Introduction}

Financial institutions hold vast repositories of transaction data that could accelerate fraud detection research, yet privacy regulations severely limit data sharing~\citep{potluru2024synthetic}. Differentially private (DP) synthetic data generation offers a promising solution, but established marginal methods like AIM~\citep{mckenna2021winning} and PrivBayes~\citep{zhang2017privbayes} face fundamental scalability challenges: utility degrades rapidly as dimensionality increases~\citep{zhang2017privbayes}, limiting their applicability to carefully engineered low-dimensional schemas.

LLM-based simulators present an intriguing alternative. Rather than learning distributions directly, they leverage pre-trained knowledge to generate realistic data from high-level descriptions. This suggests a ``Profile-then-Simulate'' approach: use DP mechanisms to synthesize compact user personas, then feed these into an LLM simulator to generate rich transaction sequences. If LLMs can faithfully reproduce the statistical properties of their DP inputs, this architecture could scale to complex user representations where direct synthesis fails.

We evaluate this hypothesis using PersonaLedger~\citep{yuan2026personaledger}, an agentic financial simulator, seeded with DP synthetic personas derived from real user statistics. To establish a meaningful baseline, we also apply direct DP synthesis to a carefully engineered 12-feature transaction schema---a best-case scenario for marginal methods. This controlled comparison isolates the question: can LLM simulators match the fidelity of direct synthesis when the problem is tractable, as a prerequisite for tackling problems where direct synthesis cannot?

\textbf{Contributions:} (1) A methodology for seeding LLM simulators with DP-protected personas derived from real user statistics; (2) controlled comparison against direct DP synthesis to benchmark LLM fidelity; and (3) identification of systematic LLM biases that currently limit this approach.

\section{Methodology: The Decoupled Framework}

Our approach uses a two-phase process: (1) DP synthesis of user personas from real behavioral statistics and (2) LLM-based transaction generation from these personas. This ``Profile-then-Simulate'' architecture could enable DP synthesis for complex user representations by offloading generation to pre-trained models, assuming LLMs can faithfully reproduce input distributions.

\subsection{Phase 1: Privacy-Preserving Persona Generation}

We aggregate raw transaction logs ($N \approx 24M$ transactions from 2,000 users) into behavioral profiles, computing 27 user-level summary statistics (Table~\ref{tab:ss_schema}) that capture demographics, spending habits, and risk factors. This user-level aggregation preserves the connection between who a user is and how they transact~\citep{ganev2025importance}, while providing an optimal input format for marginal DP synthesis methods~\citep{zhang2017privbayes}.

We discretize summary statistics prior to synthesis using domain-informed quantile binning to preserve fraud-relevant signal while optimizing for DP efficiency~\citep{mckenna2021winning}. For attributes requiring high precision (e.g., income, amount\_mean, amount\_max), we employ deciles to maintain granular distinctions. For attributes where coarser categories remain informative (age, debt, spending proportions), we use quintiles or quartiles. The fraud risk indicator (perc\_fraud\_txns) uses 7 quantiles to provide multiple risk levels with sufficient granularity for downstream classification tasks.

We generate DP synthetic personas using AIM~\citep{mckenna2021winning} (Adaptive Information Mechanism) across privacy regimes $\epsilon \in \{1, 5, 10\}$. AIM's superior performance on small tabular datasets ($N=2,000, d=27$) compared to GAN-based approaches~\citep{chen2025benchmarking} makes it optimal for maximizing utility in DP persona generation, particularly given GANs' requirement for large-scale data to achieve high-quality synthesis~\citep{jordon2018pate}.

\subsection{Phase 2: Agentic Simulation (PersonaLedger)}

PersonaLedger~\citep{yuan2026personaledger} is a rule-grounded, LLM-driven simulator that generates transaction sequences from user personas. We convert DP synthetic personas into PersonaLedger format through structured mapping that preserves privacy guarantees, transforming the 27 summary statistics into UserPersona (demographics/behavior) and UserFinancialProfile (spending/payments) components while normalizing relative spending proportions to sum to unity.

To ensure compatibility with downstream evaluation, we extend PersonaLedger with target schema generation that produces transactions in a 12-dimensional ordinal format. This includes categorical mappings for transaction amounts (10 levels), merchant codes (4 groups), payment methods (3 types), and demographics, with fraud probability derived from persona risk attributes. Category-aware prompting guides LLM generation within the constrained ordinal space, maintaining consistency between DP persona attributes and synthetic transaction patterns while preserving privacy guarantees.

This architectural separation allows the privacy budget to focus entirely on user-level statistics while the simulator ensures transactions are consistent with each persona's characteristics.

\section{Experimental Setup}

We use the Kaggle Credit Card Transactions dataset\footnote{\url{https://www.kaggle.com/datasets/ealtman2019/credit-card-transactions}}, a multi-agent simulation of 20M+ transactions from 2,000 synthetic consumers that matches real fraud data across key dimensions. We apply strict TSTR (Train-on-Synthetic, Test-on-Real) protocol with careful attention to fraud rate handling across datasets.

\textbf{Test Data:} The holdout set (20\%, 400 users, 4.8M transactions) preserves the natural fraud rate ($\sim$0.13\%). We draw a stratified sample of 100K transactions for evaluation, yielding $\sim$130 fraud cases while maintaining the natural class distribution. This low fraud prevalence motivates our use of AUC as the primary metric, as it is threshold-independent and robust to severe class imbalance.

\textbf{Training Data:} For direct synthesis baseline, we transform raw transaction logs into a transaction-level dataset (80K transactions) with oversampled fraud to create a 25\% fraud rate, preserving all fraud transactions from training users. This stratified sampling aligns with AIM's IID assumptions~\citep{mckenna2021winning}. For PersonaLedger, we generate $\sim$5K transactions per privacy regime at a near-natural fraud rate ($\sim$3\%), as the simulator's rule-based logic produces fraud patterns without requiring oversampling.

\textbf{Evaluation Protocol:} Both synthetic datasets train XGBoost classifiers evaluated on the same held-out real transactions at natural fraud rates. The strategy of training on elevated fraud rates and testing on natural rates aligns with realistic deployment scenarios where fraud-enriched synthetic data enables model development, while evaluation must reflect production conditions.

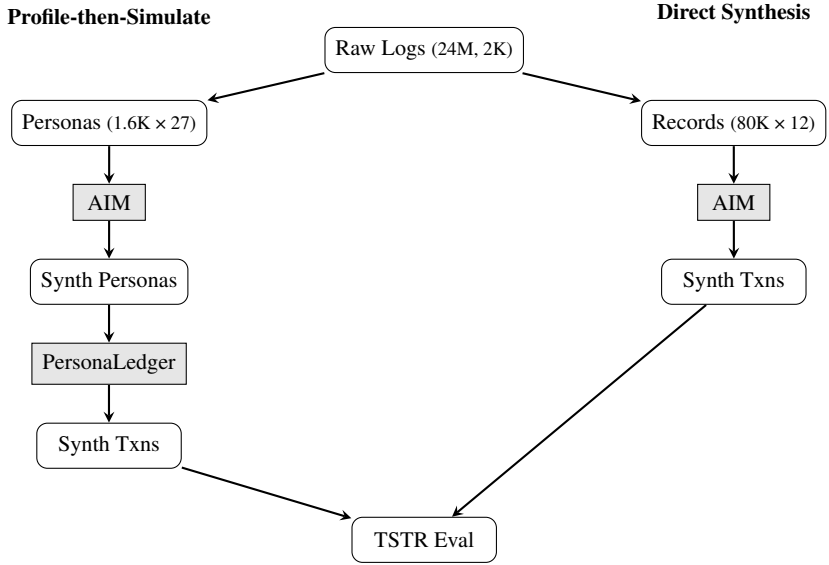
\begin{figure}[htbp]
\centering
\tikzstyle{box} = [rectangle, rounded corners, minimum width=1.6cm, minimum height=0.5cm, text centered, draw=black, fill=white, font=\scriptsize]
\tikzstyle{process} = [rectangle, minimum width=0.8cm, minimum height=0.35cm, text centered, draw=black, fill=gray!20, font=\scriptsize]
\tikzstyle{arrow} = [thick,->,>=stealth]

\begin{tikzpicture}[node distance=0.5cm, scale=1.2, every node/.style={scale=1.2}]

\node (raw_data) [box] {Raw Logs {\tiny (24M, 2K)}};

\node (personas) [box, below left=of raw_data, xshift=-1cm, yshift=0cm] {Personas {\tiny (1.6K × 27)}};
\node (transactions) [box, below right=of raw_data, xshift=1cm, yshift=0cm] {Records {\tiny (80K × 12)}};

\node (dp_personas) [process, below=of personas] {AIM};
\node (dp_transactions) [process, below=of transactions] {AIM};

\node (synth_personas) [box, below=of dp_personas] {Synth Personas};
\node (synth_transactions) [box, below=of dp_transactions] {Synth Txns};

\node (personaledger) [process, below=of synth_personas] {PersonaLedger};
\node (synth_sequences) [box, below=of personaledger] {Synth Txns};

\node (evaluation) [box, below=of raw_data, yshift=-4.5cm] {TSTR Eval};

\draw [arrow] (raw_data) -- (personas);
\draw [arrow] (raw_data) -- (transactions);
\draw [arrow] (personas) -- (dp_personas);
\draw [arrow] (transactions) -- (dp_transactions);
\draw [arrow] (dp_personas) -- (synth_personas);
\draw [arrow] (dp_transactions) -- (synth_transactions);
\draw [arrow] (synth_personas) -- (personaledger);
\draw [arrow] (personaledger) -- (synth_sequences);
\draw [arrow] (synth_sequences) -- (evaluation);
\draw [arrow] (synth_transactions) -- (evaluation);

\node [above=of personas, yshift=0.3cm, font=\scriptsize\bfseries] {Profile-then-Simulate};
\node [above=of transactions, yshift=0.3cm, font=\scriptsize\bfseries] {Direct Synthesis};

\end{tikzpicture}
\caption{Experimental setup comparing two approaches: Profile-then-Simulate (left) aggregates 24M transactions into 1.6K user personas with 27 behavioral features then uses PersonaLedger to generate synthetic transaction sequences, while Direct Synthesis (right) samples 80K transaction records with 12 attributes. Both use AIM for DP synthesis before TSTR evaluation.}
\label{fig:methodology}
\end{figure}

We compare against AIM~\citep{mckenna2021winning} (Adaptive Information Mechanism) applied to both our persona-based approach and direct synthesis on raw transaction logs using Private PGM implementations. This controlled comparison isolates the impact of our decoupling architecture while using the same DP mechanism (Figure~\ref{fig:methodology}). Evaluation spans privacy regimes $\epsilon \in \{1, 5, 10\}$ representing high-privacy stress testing to high-utility internal sharing.

We evaluate across multiple dimensions~\citep{hernandez2024standardised,steier2025synthetic}: (1) \textit{Utility}: XGBoost fraud detection AUC on real holdout data with 100 bootstrap iterations for confidence intervals, and (2) \textit{Fidelity}: Total Variation Distance (TVD) on 1-way marginals and 2-way correlations to measure how well the LLM reproduces input statistical distributions.

\vspace{-1em}
\section{Results \& Discussion}

\subsection{Utility (TSTR)}

Table~\ref{tab:results} presents fraud detection performance using AUC on held-out real transactions at natural fraud rates.

\begin{table}[htbp]
\centering
\begin{threeparttable}
\caption{TSTR Evaluation Results: AUC on real test data (100K sampled transactions, 0.13\% fraud)}
\label{tab:results}
\begin{tabular}{|l|c|c|c|c|c|}
\hline
\textbf{Method} & \textbf{$\epsilon$} & \textbf{AUC} & \textbf{95\% CI} & \textbf{1-way TVD} & \textbf{2-way TVD} \\
\hline
PersonaLedger & 1 & 0.70 & (0.64, 0.74) & 0.32 & 0.48 \\
PersonaLedger & 5 & 0.51 & (0.46, 0.56) & 0.30 & 0.47 \\
PersonaLedger & 10 & 0.57 & (0.52, 0.62) & 0.34 & 0.52 \\
\hline
AIM Synthetic & 1 & 0.87 & (0.85, 0.90) & 0.05 & 0.10 \\
AIM Synthetic & 5 & 0.89 & (0.86, 0.91) & 0.05 & 0.09 \\
AIM Synthetic & 10 & 0.89 & (0.86, 0.91) & 0.05 & 0.09 \\
\hline
\end{tabular}
\end{threeparttable}
\end{table}

PersonaLedger achieves moderate utility at $\epsilon=1$ (AUC 0.70), demonstrating that LLM-based generation from DP statistics can produce usable synthetic data. Notably, the non-monotonic relationship with privacy budget ($\epsilon=5$ performs worst) suggests that LLM generation biases, rather than DP noise, dominate performance variation.

\subsection{Fidelity Analysis}

PersonaLedger exhibits substantial distribution drift (TVD 0.30-0.34). Feature-level analysis reveals systematic LLM biases:

\begin{itemize}
\item \textbf{Temporal features:} Transactions cluster in ``business hours'' (9am-2pm), missing night/evening patterns present in real data
\item \textbf{Demographics:} Generated personas skew older and disproportionately retired ($\sim$80\% vs $\sim$50\% in real data)
\item \textbf{Well-preserved features:} Binary/low-cardinality attributes (gender, home\_zip\_txn, day\_of\_week) transfer accurately
\end{itemize}

These biases reflect LLM priors about ``typical'' financial behavior overriding the statistical distributions provided as input.

\subsection{Limitations}

Our findings reveal a fundamental tension in LLM-based synthetic data generation: the model's learned priors about realistic behavior can conflict with target statistical distributions. Features aligned with common-sense expectations transfer well, while counter-intuitive patterns (late-night transactions, young high-spenders) exhibit drift. Additionally, the current pipeline does not fully propagate all DP statistics through to the final output schema, requiring post-hoc estimation for some attributes.

\section{Conclusion}

We demonstrate that LLM-based simulators can generate transactions from DP-protected real user statistics with promising utility (AUC 0.70 at $\epsilon=1$). However, systematic LLM biases cause significant distribution drift, particularly for temporal and demographic features where learned priors override input statistics. These findings suggest that techniques for enforcing stricter adherence to input distributions are needed before LLM-based methods can fulfill their promise for complex high-dimensional user data.

\textbf{Future Work:} Constrained decoding or rejection sampling to enforce distribution adherence; evaluation on high-dimensional user profiles where direct synthesis is infeasible; and prompt engineering strategies to reduce LLM biases for financial data.

\bibliography{iclr2026_conference}
\bibliographystyle{iclr2026_conference}

\newpage
\appendix
\section{Data Schema}

\vspace{-1em}
\begin{table}[H]
\centering
\caption{Summary Statistics Schema: 27 features extracted from transaction logs}
\label{tab:ss_schema}
\begin{tabular}{|l|l|l|p{5.5cm}|}
\hline
\textbf{Column Name} & \textbf{Type} & \textbf{Bins} & \textbf{Description} \\
\hline
is\_high\_risk & binary & 2 & 1 if User is in top 15\% of perc\_fraud\_txns \\
age\_years & int & 5 & Current Age in years \\
is\_retired & binary & 2 & 1 if Current Age $\geq$ Retirement Age \\
gender & binary & 2 & 1 for female; 0 for male \\
zip\_income\_per\_capita & float & 5 & Mean income per capita in home zipcode \\
income & float & 10 & Total income amount \\
debt & float & 5 & Total debt amount \\
n\_cards & int & 5 & Number of cards \\
amount\_total & float & 10 & Total spend across all transactions \\
amount\_mean & float & 10 & Mean transaction amount \\
amount\_max & float & 10 & Largest single transaction \\
amount\_std & float & 10 & Std dev of transaction amounts \\
mean\_daily\_txns & float & 5 & Mean number of daily transactions \\
max\_daily\_txns & int & 5 & Max number of daily transactions \\
mean\_monthly\_txns & float & 5 & Mean number of monthly transactions \\
max\_monthly\_txns & int & 5 & Max number of monthly transactions \\
prop\_chip\_txns & float & 5 & Proportion of chip transactions \\
prop\_online\_txns & float & 5 & Proportion of online transactions \\
prop\_swipe\_txns & float & 5 & Proportion of swipe transactions \\
n\_unique\_mcc & int & 5 & Number of unique MCCs \\
prop\_spend\_food & float & 5 & Proportion spend in Retail-Food \\
prop\_spend\_goods & float & 5 & Proportion spend in Retail-Goods \\
prop\_spend\_services\_other & float & 5 & Proportion spend in Services-Other \\
prop\_spend\_travel & float & 5 & Proportion spend in Travel \\
prop\_homezip\_txns & float & 5 & Proportion txns in home zipcode \\
prop\_nozip\_txns & float & 5 & Proportion txns with no zipcode \\
perc\_fraud\_txns & float & 7 & Percent of fraud transactions \\
\hline
\end{tabular}
\end{table}

\vspace{-0.5em}
\begin{table}[H]
\centering
\caption{Transaction Schema: 12 features used for direct synthesis and evaluation}
\label{tab:txn_schema}
\begin{tabular}{|l|l|l|p{5.5cm}|}
\hline
\textbf{Column Name} & \textbf{Type} & \textbf{Bins} & \textbf{Description} \\
\hline
is\_fraud & binary & 2 & 1 if transaction is fraudulent; 0 otherwise \\
amount & ordinal & 10 & Transaction amount (decile bins) \\
txn\_method & categorical & 3 & Payment method: Chip, Swipe, Online \\
day\_of\_week & ordinal & 7 & Day of week (0=Monday, 6=Sunday) \\
hour & ordinal & 4 & Time of day (Night, Morning, Afternoon, Evening) \\
mcc\_group & categorical & 4 & Merchant category: Travel, Retail\_Food, Retail\_Goods, Services\_Other \\
home\_zip\_txn & binary & 2 & 1 if transaction in user's home zipcode \\
age\_years & ordinal & 5 & User age (quintile bins) \\
gender & binary & 2 & 1 for female; 0 for male \\
is\_retired & binary & 2 & 1 if user is retired \\
income & ordinal & 5 & User income (quintile bins) \\
zip\_income\_per\_capita & ordinal & 5 & Zipcode income per capita (quintile bins) \\
\hline
\end{tabular}
\end{table}

\end{document}